\documentclass[letterpaper, 10 pt, conference]{ieeeconf}  

\IEEEoverridecommandlockouts                              

\usepackage{graphics} 
\usepackage{epsfig} 
\usepackage{mathptmx} 
\usepackage{times} 
\usepackage{amsmath} 
\usepackage{amssymb}  
\usepackage{multirow}
\usepackage{graphicx}
\usepackage{caption}
\usepackage{cite}
\usepackage{hyperref} 
\usepackage{color}

\usepackage{pifont}
\newcommand{\cmark}{\ding{51}}%
\newcommand{\xmark}{\ding{55}}%

\title{\LARGE \bf
ImitationNet: Unsupervised Human-to-Robot Motion Retargeting via Shared Latent Space
}

\author{Yashuai Yan\footnotemark$^{*1}$, Esteve Valls Mascaro\footnotemark$^{*1}$, and Dongheui Lee$^{1,2}$
\thanks{$^{1}$Yashuai Yan and Esteve Valls Mascaro and Dongheui Lee are with Autonomous Systems, Technische Universität Wien (TU Wien), Vienna, Austria (e-mail: \texttt{\{yashuai.yan,esteve.valls.mascaro, dongheui.lee\}@tuwien.ac.at}).}%
\thanks{$^{2}$Dongheui Lee is also with the Institute of Robotics and Mechatronics (DLR), German Aerospace Center, Wessling, Germany.}\\%
\href{https://evm7.github.io/UnsH2R/}{\color{blue} evm7.github.io/UnsH2R}
}

\begin{document}

\thispagestyle{empty}
\pagestyle{empty}

\twocolumn[{%
\renewcommand\twocolumn[1][]{#1}%
\maketitle
\begin{center}
    \centering
    \vspace{-6mm}
    \captionsetup{type=figure}
    \includegraphics[width=.98\textwidth]{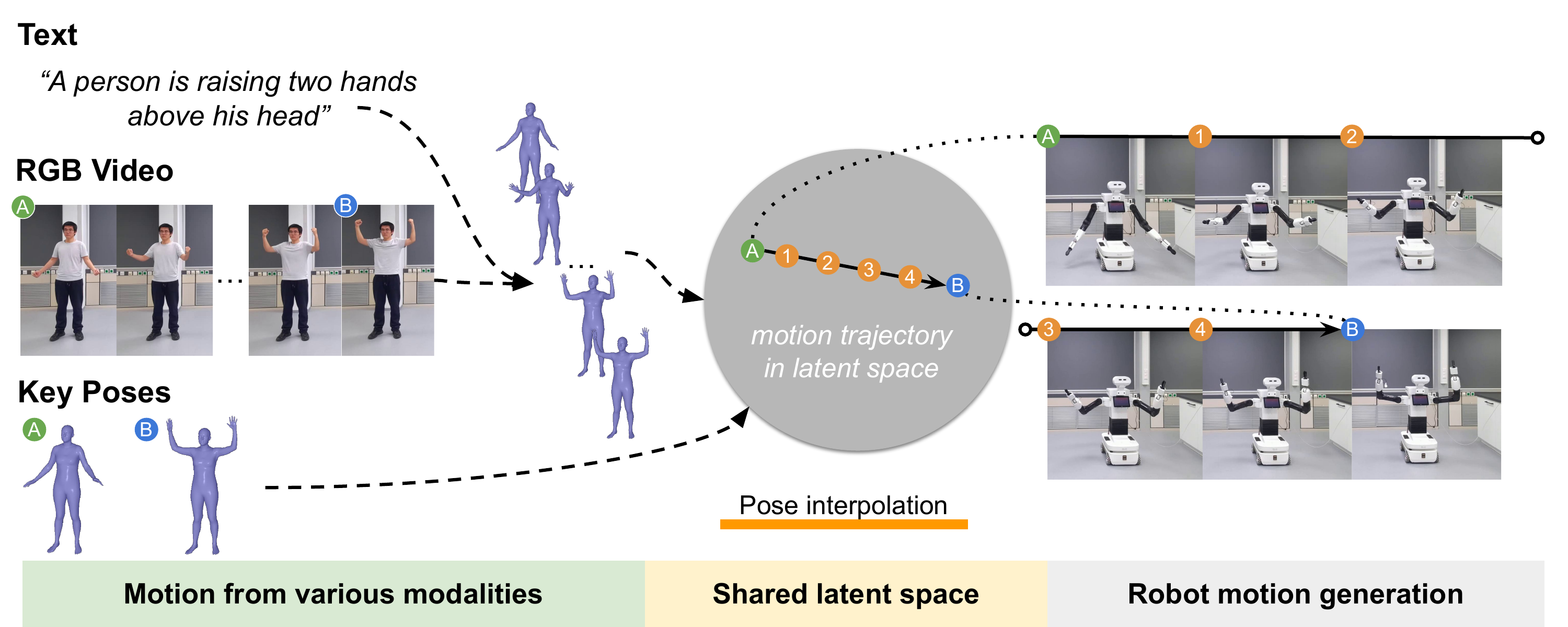}
    \captionof{figure}{Our human-to-robot motion retargeting connects robot control with diverse source modalities, such as a text description, an RGB video, or key poses. Our approach can encode human skeletons into a  shared latent space between humans and robots, and subsequently decode these latent variables into the robot's joint space, enabling direct robot control. Additionally, our approach facilitates the generation of smooth robot motions between human key poses (represented as green and blue dots) through interpolation within the latent space (indicated by the orange dots).}
\end{center}%
}]

\renewcommand{\thefootnote}{\fnsymbol{footnote}}
\footnotetext{$^{1}$Yashuai Yan,  Esteve Valls Mascaro, and Dongheui Lee are with Autonomous Systems Lab, Technische Universität Wien (TU Wien), Vienna, Austria (e-mail: \texttt{\{yashuai.yan, esteve.valls.mascaro, dongheui.lee\}@tuwien.ac.at}).}

\footnotetext{$^{2}$Dongheui Lee is also with the Institute of Robotics and Mechatronics (DLR), German Aerospace Center, Wessling, Germany.}
\footnotetext{$^{*}$Contributed equally to the work.}
\begin{abstract}

This paper introduces a novel deep-learning approach for human-to-robot motion retargeting, enabling robots to mimic human poses accurately. Contrary to prior deep-learning-based works, our method does not require paired human-to-robot data, which facilitates its translation to new robots. First, we construct a shared latent space between humans and robots via adaptive contrastive learning that takes advantage of a proposed cross-domain similarity metric between the human and robot poses.  Additionally, we propose a consistency term to build a common latent space that captures the similarity of the poses with precision while allowing direct robot motion control from the latent space. For instance, we can generate in-between motion through simple linear interpolation between two projected human poses. We conduct a comprehensive evaluation of robot control from diverse modalities (i.e., texts, RGB videos, and key poses), which facilitates robot control for non-expert users. Our model outperforms existing works regarding human-to-robot retargeting in terms of efficiency and precision. Finally, we implemented our method in a real robot with self-collision avoidance through a whole-body controller to showcase the effectiveness of our approach.

\end{abstract}

\section{INTRODUCTION}

In recent years, human-robot interaction (HRI) has gained significant attention as it plays a leading role in deploying robots into our daily lives. For a natural HRI, the robot needs not only to capture the human movements but also to understand the human motion intentions behind them. To enhance HRI, it is also crucial to intuitively retarget these human motions onto robots while preserving their similarity and improving robot autonomy. This paper addresses the challenge of enabling robots to mimic human motions while preserving the likeness of the original movement. 

However, retargeting human motions to robots is a complex task due to the fundamental differences between human and robot anatomies, kinematics, and motion dynamics. Unlike humans, robots possess rigid bodies, different form factors, and distinct physical limitations. Consequently, directly mapping human motion to robot actuators often leads to unnatural and suboptimal robot behavior, undermining the objective of achieving human-like movements. For example, when retargeting the motion of a human touching his head with his right hand, it is crucial that the retargeted robot poses also reproduce this touching behavior in their motion. Solely replicating the specific arm movements could lead to the robot hand not being close to the head due to the different robot kinematics. Encoding such motions in the retargeting task is essential to ensure that robots have more natural and intuitive behaviors, leading to better and easier HRI.

While motion retargeting is a long-standing challenge in the robotic and animation community, most recent research has been focused on the exploration of large human motion capture datasets \cite{human36m, HumanML3D} to learn and synthesize human motions from different modality inputs: text \cite{mdm}, 3D scene \cite{scenemotion}, audio\cite{audiomotion} or conditioned by key poses \cite{2CH-TR}. Our primary goal in this research is to develop a novel method that eliminates the reliance on data annotation, thereby accomplishing the learning of a shared representation space in which human and robot poses are mutually and integrally represented. A good representation space ensures that similar poses from both domains are positioned close to each other while dissimilar poses are far apart. While previous research \cite{rt_pair_data_1, rt_pair_data_2} requires manually annotating human and robot pairs performing the same pose to learn this retargeting process, we consider an unsupervised training technique that does not require pairing data. Consequently, we can reduce the implementation costs for retargeting human poses to new robots. 

To this end, we propose an encoder-decoder architecture to construct a latent space that preserves the spatial relationships between human joints as well as the likeness of the original human motion. We achieve this process through the synergy of multiple losses. First, we adopt adaptive contrastive learning to autonomously construct the common latent space based on a proposed similarity metric. Then, we incorporate a reconstruction loss on robot data to ensure the regeneration of the same motion from the latent space.   that the robot faithfully follows the movement of the human. Finally, we enforce a consistency term to constrain that the robot faithfully follows the movement of the human. As a consequence, the constructed latent space remains tractable via simple operations. For instance, we are able to generate smooth robot motions between key poses by simply using linear interpolation in the latent space. This intuitive behavior facilitates motion control and also showcases the robustness of our learned latent space. Finally, our decoder can translate the latent representations to robot motion control commands. Contrary to prior methods that adopt soft safety measures in learned approach \cite{rt_selfsupervised_hyemin}, we implement our method in a real robot with a whole-body controller that ensures self-collision avoidance in the retargeted motion. 

Our pipeline allows for the seamless and real-time translation of human skeleton data into robot motion control. Additionally, our model can be easily integrated into the aforementioned deep learning architectures \cite{mdm, scenemotion, audiomotion} to accommodate robot motion control from various modalities, enabling flexible and intuitive control over robot behavior. By addressing this challenge, we anticipate significant advancements in HRI. Our research has broad applications, including robot-assisted therapy, entertainment, teleoperation, and industrial robotics. Enabling robots to replicate human motion and intention opens up new possibilities for intuitive and natural HRI, enhancing user experience and fostering acceptance of integrating robots into our daily lives. Our work leads to the following contributions.

\begin{enumerate}
    \item Unsupervised deep learning approach to learn human-to-robot retargeting without any paired human and robot motion data.
    \item Robust and tractable latent space to generate smooth robot motion control through simple linear interpolation.
    \item Direct mapping from human skeletons to robot control commands via an encoder-decoder neural network.
    \item Evaluating control of a real robot from various modalities: text, video, or conditioned by key poses, which ensures user-friendly robot control, particularly for non-experts.
    
\end{enumerate}

\section{RELATED WORK}
Existing literature on human-to-robot motion retargeting techniques is reviewed next, highlighting limitations and the need for advancements in translating human motion's overall expressivity and naturality.

\subsection{Motion retargeting in animation}
Human motion retargeting onto animated characters has been a long-standing challenge in the computer graphics community. By bridging the gap between human motion and animation, motion retargeting enhances the quality and naturality of character animation, opening up possibilities for various applications in fields such as film, gaming, and virtual reality.

Classical motion retargeting approaches \cite{ANI_RET_IK_1, ANI_RET_IK_2, Lee, Dongheui} involved manually defining kinematic constraints and simplifying assumptions to map human motion onto animated characters. These methods were limited in their ability to handle complex motions and could not accurately capture human movement's nuances. However, with the increased availability of motion capture data \cite{human36m, HumanML3D}, data-driven approaches emerged as a more attractive alternative. These approaches offer the potential to overcome the limitations of classical methods and achieve more natural and nuanced motion transfer. \cite{rt_pair_data_1, rt_pair_data_2} learned a shared latent representation to translate motions between different kinematic agents. However, they required paired training data, which is costly and specific for each robot. To cope with the cost of pairing data, \cite{rt_anim_firstdeep} used a recurrent neural network to learn motion retargeting without those pairs using adversarial training and cycle consistency. \cite{rt_anim_pmnet} showed that disentangling pose from movement in the retargeting process leads to more natural outcomes. However, these data-driven approaches required the same source and target kinematics. Inspired by the intuition that different kinematics can be reduced to a common primal skeleton, \cite{rt_animation_skeletonaware} proposed explicitly encoding the different skeleton topologies and projecting those into a shared latent space without pairing data. \cite{rt_animation_skeletonaware} adopted a latent consistency loss to ensure that the retargeted poses remain faithful to the source. Our work is inspired by their consistency idea, but we construct a more robust shared latent space through a contrastive loss which improves the retargeting outcome. Recently, \cite{rt_animation_contactware, rt_animation_dist_matrix} focused on the motion retargeting but considering the mesh constraints of the animated characters, and thus adjusting motions to reduce interpenetration and feasibility of the motions. Contrary to the aforementioned works that consider self-collision avoidance as an additional feature for more realistic animation, our work ensures the feasibility of the retarget motion by implementing self-collision in the whole-body control of a real robot while preserving the source motion likeness. Finally, \cite{rt_animation_dist_matrix} proposed an Euclidean distance matrix to account for the motion retargeting, which is relevant for skeletons with similar proportions but underperforms when the targets have different trunk-to-arms ratios, as in our case. On the contrary, we propose to formulate this similarity through global rotations, which precisely capture the likeness in the retargeting task.

\subsection{Motion retargeting in robotics}
Despite the great success of motion retargeting for character animation, their community has only been considering the feasibility of the movements in terms of physical constraints \cite{rt_animation_contactware, rt_animation_dist_matrix, Lee, Kai_Hu}. Besides ensuring motion's feasibility, robotics research also requires adequate control of the appropriate robot based on the source motion. \cite{roboticretargetting, Kai_Hu} considered constrained optimization algorithms to retarget a human motion in a simulated robot but required learning a given trajectory and can not quickly overcome new variations. \cite{rt_control_humanoid} proposed Bayesian optimization and inverse kinematics (IK) to tackle natural retargeting, but their approach required manually selecting joints of interest and was constrained to a few specific motions. Likewise, \cite{robot_retarget3, robot_retarget1, Kai_Hu} considered whole-body retargeting by mapping human link orientation to robots and solving IK. \cite{robot_retarget3} introduced a dynamic filter to enforce robot stability, which also over-smoothed the robot poses, thus failing to capture the motion nuances in the retargeting. Moreover, \cite{robot_retarget3} method did not generalize to new kinematics. To cope with that issue, \cite{robot_retarget1} proposed to solve the IK over the robot model, which facilitated the generalization to new robots. For that,  \cite{robot_retarget1} orients the robot links closer to the corresponding human links to better capture the likeness in the retargeting. We adopt a similar approach by considering the global rotation of body links as the similarity measurement between humans and the retargeted robot pose. However, all these previous works failed to overcome the manual morphing problem \cite{darvish2023teleoperation}: the challenge of mapping in the joint space from human to robot, which requires similar joint orders among the human and robot. On the contrary, our work does not focus on the task of retargeting the poses while keeping the robot balanced, \cite{robot_retarget3, robot_retarget1}, but on the generalization of a unique method for human-robot retargeting with accuracy and capturing the nuances. Closer to our work, \cite{robot_retarget2} proposed a learned-based footstep planner and a whole-body controller to retarget the human locomotion to a robot while being coherent with the generated footsteps. However, \cite{robot_retarget2} only considered locomotion retargeting and assumed that the robot had at least one known contact with the environment at any time. Therefore,  \cite{robot_retarget2} was inappropriate for contact-free motions such as jumping or running.  

Deep learning has become a solution to ensure the retargeting process generalizes in terms of kinematics and diversity in the motions while being efficient. First, \cite{rt_shared_space_humanoid} proposed to construct a shared latent space to retarget human motion to humanoid robots, and the shared latent space is constructed with annotated human-to-robot pair data. Gathering a sufficient quantity of paired data for constructing the latent space is a laborious and time-intensive process and hardens the generalization to new configurations. \cite{rt_selfsupervised_hyemin} extended this approach by creating an automated paired data generation process. However, both works have to use nonparametric optimization in the latent space to retrieve similar robot poses to control the robot, which is inefficient if the dataset to retrieve is large. Contrastingly, our method learns a direct mapping from human poses to robot control commands. Therefore, our approach can control a robot at a high rate without being constrained by the quantity of training data.

\section{METHODOLOGY}
\begin{figure*}[]
    \centering
    \includegraphics[width=0.98\textwidth]{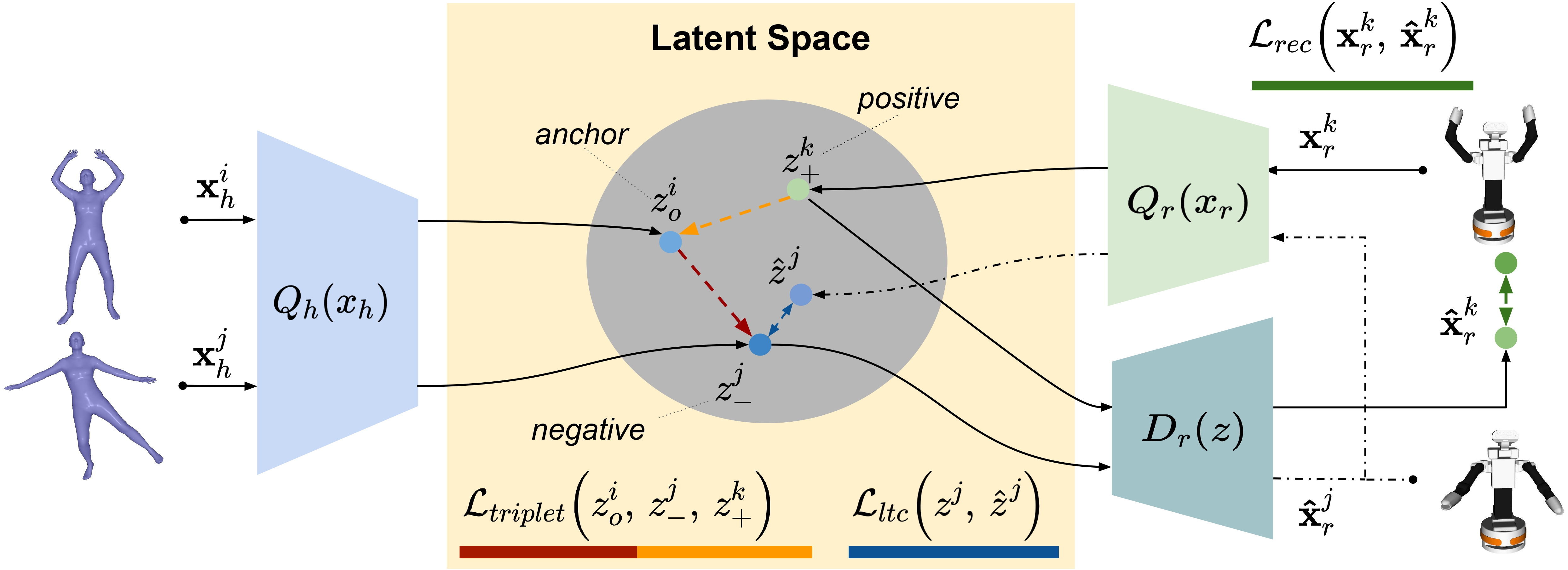}
    \caption{\textbf{Model overview.} Two human poses $(\mathbf{x}_{h}^{i}, \mathbf{x}_{h}^{j})$ are encoded into latent variables $(z^{i}, z^{j})$ within the shared space using the function $Q_{h}$. Similarly, a robot data $\mathbf{x}_{r}^{k}$ is mapped into $z^{k}$ by $Q_{r}$. Given three samples $(z^{i}, z^{j}, z^{k})$, $z^{i}$ is randomly chosen as an anchor $z_{o}^{i}$, and $z^{j}, z^{k}$ are estimated as a negative $z_{-}^{j}$ and positive $z_{+}^{k}$ sample through similarity metric in Equation \ref{eq:globaldisstancerotation}. The triplet loss $\mathcal{L}_{triplet}$ constrains the construction of the latent space by bringing $z_{o}^{i}$ and $z_{+}^{k}$ closer and pushing $z_{o}^{i}$ and $z_{-}^{j}$ apart. The decoder $D_{r}$ decodes latent variable $z^{k}$ into $\mathbf{\hat{x}}_{r}^{k}$ that should be consistent with the robot data $\mathbf{x}_{r}^{k}$ regarding $\mathcal{L}_{rec}$. The latent variable $z^{j}$ from the human data $\mathbf{x}_{h}^{j}$ is mapped into a robot data $\mathbf{\hat{x}}_{r}^{j}$. To ensure that $\mathbf{\hat{x}}_{r}^{j}$ is from the same distribution as $\mathbf{x}_{r}^{k}$, $Q_{r}$ encodes $\mathbf{\hat{x}}_{r}^{j}$ back to latent variable $\hat{z}^{j}$, and $\mathcal{L}_{ltc}$ minimizes the distance between  $\hat{z}^{j}$ and $z^{j}$. During the inference phase, $\mathbf{\hat{x}}_{r}^{j}$ is used to control the robot directly to mimic human pose $\mathbf{x}_{h}^{j}$.}
    \label{fig:modeloverview}
\end{figure*}

In this section, we present an overview of our proposed framework for unsupervised human-to-robot motion retargeting via a shared latent space. First, we formulate the human-to-robot retargeting task. Then, we describe our encoder-decoder deep learning architecture, illustrated in Figure \ref{fig:modeloverview}.

\subsection{Problem Formulation}

Let $\mathbf{x}_{h}=[x_{h,1},\cdots,x_{h,J_h}]  \in \mathbb{R}^{J_h\times n}$ be a human pose composed by $J_h$ joints. Similarly, $\mathbf{x}_{r}=[x_{r, 1}, \cdots, x_{r, J_r}]  \in \mathbb{R}^{J_r \times s}$ represents a robot pose. Then, the task of human motion retargeting can be formulated as finding a function $f$ that maps a $\mathbf{x}_{h}$ to $\mathbf{x}_{r}$ ($f: \mathbf{x}_{h} \longmapsto \mathbf{x}_{r}$) so that $\mathbf{x}_{r}$ preserves the human-like naturality of the pose $\mathbf{x}_{h}$. However, the joints for humans and robots usually have different configurations: a human joint (e.g., wrist joint) can have more than 1DoF, while one robot joint usually has only 1DoF. To cope with such differences, we describe each human joint $x_{h,j}$ as its quaternion representation referring to its parent ($n=4$),  while each robot joint $x_{r, j}$ (i.e., revolute joint) is described as its joint angle ($s=1$). 

In our particular case, and contrary to all works focusing on character animation, we are interested in the direct control of a robot. Robots can be controlled via their joint angles. As joint angles for robots and humans have different configurations, it makes little sense to compare joint angles to measure their similarity. Inspired by \cite{robot_retarget1}, we propose to use the global rotation of body links to compare the similarity between human and robot poses, which better captures their likeness and allows for better generalization to different kinematics. The similarity metric is defined in Section \ref{subsec:similarity-metric}.

Previous works \cite{rt_shared_space_humanoid, rt_selfsupervised_hyemin} rely on the acquisition of a dataset of mapped motions between the human and the robot to retarget, which we describe as a $\{\mathbf{x}_{h}, \mathbf{x}_{r}\}$ pair. These works learn the retargeting function $f$ in a supervised manner. On the contrary, we consider the retargeting task without collecting the correct $\{\mathbf{x}_{h}, \mathbf{x}_{r} \}$ pair and learn without supervision how to approximate $f$ better. To this end, our model first learns to project human $\mathbf{x}_{h}$ and robot $\mathbf{x}_{r}$ poses to the same representation space. Then, we decode the learned representation to robot joint angles, which allows us to control the robot directly.

\subsection{Cross-domain similarity metric}
\label{subsec:similarity-metric}
To create a shared latent space in an unsupervised way, we initially define a similarity metric that captures the likeness of the poses between humans and robots. Contrary to prior works that use the local quaternions \cite{rt_animation_skeletonaware} or the relative XYZ position of the end effector \cite{rt_animation_dist_matrix}, we consider the global rotation of body limbs as the similarity metric that better preserves the skeleton visual appearance. By using global rotation, our model captures the complete 3D orientation and remains invariant to coordinate systems and articulation variations. Let $q_{h, j}$ and $q_{r, j}$ represent the global quaternions of the same limbs (e.g., shoulder-to-elbow, elbow-to-wrist, etc.) of a human pose $\mathbf{x}_{h}$ and a robot pose $\mathbf{x}_{r}$. As a human pose is represented as limb quaternions, it is straightforward to obtain $q_{h, j}$ from $\mathbf{x}_{h}$. To get limb quaternions of a robot, we utilize forward kinematics to map robot joints $\mathbf{x}_{r}$ to its limb quaternions $q_{r, j}$. Then, the distance between the two poses can be computed as shown in Equation \ref{eq:globaldisstancerotation}, where $<,>$ denotes the dot product between two vectors.
\begin{equation}
     S_{GR}(\mathbf{x}_{h}, \mathbf{x}_{r}) = \sum_{j} (1- <q_{h, j}, q_{r, j}> ^2)
\label{eq:globaldisstancerotation}
\end{equation}

$S_{GR}$ is employed to measure the similarity between two poses used for contrastive learning in Section \ref{Human-to-Robot-shared-representation}.

\subsection{Human-to-Robot shared representation}
\label{Human-to-Robot-shared-representation}
We formulate the task of motion retargeting as the translation between two domains. We adopt two multi-layer perceptron (MLP) encoders ($Q_{h}$, $Q_{r}$) to project the human and robot poses to a shared representation space, respectively. This way, $Q_{h}$ projects   $\mathbf{x}_{h} \in \mathbb{R}^{J_h\times n}$ to  $z \in \mathbb{R}^{d}$ while  $Q_{r}$ translates $\mathbf{x}_{r} \in \mathbb{R}^{J_r\times s}$ to  $z \in \mathbb{R}^{d}$. Given a human pose $\mathbf{x}_{h}$, our shared latent space is used as a bridge to generate $\mathbf{x}_{r}$ while conserving its similarity defined in Section \ref{subsec:similarity-metric}.

We propose to learn the retargeting function $f: \mathbf{x}_h \longmapsto \mathbf{x}_r$ without any paired human and robot motion data. Inspired by the recent success of contrastive learning methods (e.g., CLIP \cite{radford2021learning}), we propose to construct a shared latent space between two domains (here human and robot poses) in an unsupervised manner. Contrastive learning is a training technique that aims to learn from unlabeled data by comparing and contrasting different instances according to given similarity metrics. To do that, a neural network is optimized to maximize the agreement between positive pairs (similar instances) and minimize the agreement between negative pairs (dissimilar instances).

Let us assume a large set of data that contains feasible human poses $\mathbf{x}_{h}$ and robot poses $\mathbf{x}_{r}$. Our method randomly selects triplets of projections from these data instances. As shown in Figure \ref{fig:modeloverview}, $\mathbf{x}_{h}^{i}, \mathbf{x}_{h}^{j}$ and $\mathbf{x}_{r}^{k}$ are a triplet. Then, we first encode them to the shared latent space through $Q_{h}$ and $Q_{r}$, respectively. For the encoded triplet ($z^{i}, z^{j}, z^{k}$), $z^{i}$ is randomly selected as an anchor $z_{o}^{i}$, which serves as the reference. We compute the global rotation distance $S_{GR}$ detailed in Equation \ref{eq:globaldisstancerotation} to obtain the similarity between our anchor pose $z_{o}^{i}$ and the two other poses ($z^{j}, z^{k}$). The dissimilar $z^{j}$ is a negative sample $z_{-}^{j}$ while $z^{k}$ is a positive sample $z_{+}^{k}$.

Then, we adopt the Triplet Loss \cite{hoffer2018deep} that pulls similar samples (anchor $z_{o}^{i}$ and positive $z_{+}^{k}$) close while simultaneously pushing dissimilar samples (anchor $z_{o}^{i}$ and negative $z_{-}^{j}$) away in the latent space. This allows a representation space where similar instances are clustered together and dissimilar instances are pushed apart. Equation \ref{eq:tripletloss} shows the Triplet Loss $\mathcal{L}_{triplet}$ used in our scenario, where $\alpha = 0.05$.

\begin{equation}
     \mathcal{L}_{triplet} = max(||z_{o}^{i} - z_{+}^{k}||_{2} - ||z_{o}^{i} - z_{-}^{j}||_{2} + \alpha, 0)
\label{eq:tripletloss}
\end{equation}

\subsection{Shared representation to robot control}
Our proposed encoders allow us to project human poses and robot poses into a shared representation space. Therefore, the next step is to learn how to decode latent variables $\mathbf{z}$ sampled from the shared space into robot joint space that can be directly used to control the robot. As shown in Figure \ref{fig:modeloverview}, the decoder $D_{r}$ decodes the latent variables $z^{j}$ and $z^{k}$ to robot data $\mathbf{\hat{x}}_{r}^{j}$ and $\mathbf{\hat{x}}_{r}^{k}$, respectively. As $z^{k}$ is encoded from the robot data $\mathbf{x}_{r}^{k}$, we employ a standard reconstruction loss over $\mathbf{\hat{x}}_{r}^{k}$ and $\mathbf{x}_{r}^{k}$, as shown in Equation \ref{eq:rec}. Additionally, to ensure that the predicted robot data $\mathbf{\hat{x}}_{r}^{j}$ from human data $\mathbf{x}_{h}^{j}$ is from the same distribution as the real robot data, we adopt the latent consistent loss shown in Equation \ref{eq:ltc} to encourage direct mapping in the retargeting process, similar to \cite{rt_animation_skeletonaware}.

\begin{equation}
     \mathcal{L}_{rec} = ||\mathbf{x_{r}} - D_{r}(Q_{r}(\mathbf{x_r}))||_{1}
\label{eq:rec}
\end{equation}

\begin{equation}
     \mathcal{L}_{ltc} = ||Q_{h}(\mathbf{x_{h}}) - Q_{r}(D_{r}(Q_{h}(\mathbf{x_h})))||_{1}
\label{eq:ltc}
\end{equation} 

Our approach employs an end-to-end training strategy, enabling the encoder to learn a shared representation space for both human and robot poses unsupervised while ensuring that this representation space is reconstructible to robot control through our decoder. The total loss employed during training is a weighing sum as described in Equation \ref{eq:lossall}, where $\lambda_{triplet} = 10, \lambda_{rec} = 5$.

\begin{equation}
    \mathcal{L} = \lambda_{triplet} \mathcal{L}_{triplet} + \lambda_{rec}\mathcal{L}_{rec} + \mathcal{L}_{ltc}
\label{eq:lossall}
\end{equation}

\section{EXPERIMENTS}
The experimental setup and datasets used to evaluate the performance of our model are presented, along with the metrics and benchmarks employed to assess the accuracy and fidelity of the retargeted robot motions.

\subsection{Experiment Settings}
The hyperparameter configurations used in our framework are listed in this subsection. The network consisting of two encoders and one decoder is trained end-to-end with a learning rate of $0.001$ and batch size of $256$. The encoder and decoder are Multi-Layer Perceptrons with the same structure; they have $6$ hidden layers, each with $128$ units. The shared latent space is of 8 dimensions. Adam \cite{kingma2014adam}, a momentum-based method, is utilized to optimize the loss function during training. We trained our model for 2.5 hours until the losses reached convergence. We did not experiment with the hyperparameters but chose default values to simplify the training. We acknowledge that further finetuning of those parameters could result in improvements in our results. We use a Ubuntu 22.04 and RTX A4000 Graphic card for our experiment. 

Additionally, we employ a bi-manual TiaGo++ robot that integrates two 7-DoF arms. In this paper, we focus on the motion of the upper and lower arm parts. We ignore the motion of the two hands because the HumanML3D human motion dataset \cite{HumanML3D} used does not contain hand motions. Therefore, the similarity metric $S_{RD}$ in Equation \ref{eq:globaldisstancerotation} is defined on four limbs: left shoulder-to-elbow, left elbow-to-wrist, right shoulder-to-elbow, and right elbow-to-wrist. 

To control the robot in the real world, we send joint commands to the whole-body-controller \cite{5174677} integrated in Tiago++ robot. The whole-body controller handles joint angle limits, joint velocity limits, and self-collision avoidance.

\subsection{Data collection}
We present a robot pose generation procedure that requires only the robot's kinematic information. First, we sample the robot joint angles from its configuration space. The robot pose can be computed by following its forward kinematics. In such a way, we collect around 15M poses from the TiaGo++ robot by randomly sampling angles per joint. For human motions, we use the HumanML3D dataset \cite{HumanML3D} that consists of 14616 motions with a total length of 28.59 hours, summing up to around 20M poses. HumanMl3D covers human daily activities (e.g., 'walking', 'jumping'), sports (e.g., 'playing golf'), acrobatics (e.g., 'cartwheel'), and artistry (e.g., 'dancing'). In HumanML3D, a human pose is represented by its skeletons. As robot poses are sampled randomly from the configuration space, they are not matched to human poses in HumanML3D.

\subsection{Baseline}
\label{sec:baseline}
We implement S$^{3}$LE \cite{rt_selfsupervised_hyemin} as our baseline. To train S$^{3}$LE, we use a similar method as mentioned in \cite{rt_selfsupervised_hyemin} to generate paired data. We generate the same amount of paired data as in \cite{rt_selfsupervised_hyemin}, 200K, by selecting the pairs with minimal rotation distance measured by Equation \ref{eq:globaldisstancerotation}. The paired data is only used to train our baseline method.

\subsection{Quantitative evaluation}
To evaluate the performance of each retargeting method, we annotated 11 distinct motions that were not observed while training. The annotated motions serve as the ground truth for our evaluation. We employ the Mean Square Error (MSE) of joint angles between ground truth and predicted results to quantify our proposed method. Furthermore, our method endeavors to address motion retargeting in real-time scenarios. We thoroughly evaluated the computational efficiency and speed at which our model operates.

Table \ref{table:quantitative_eval} compares our method with the baseline in Section \ref{sec:baseline}. Our method outperforms the baseline in terms of MSE of joint angles. Furthermore, our novel approach demonstrates a notable increase in operational efficiency, surpassing the baseline by more than a factor of three. With a speed of 1.5kHz, our method readily fulfills the requirements of most advanced robot control systems.

\begin{table}[]
\centering
\caption{Performance of our proposed method and the baseline. The Mean Square Error (MSE) of joint angles between ground truth and predicted results are compared here. Bold fonts indicate better results.}
\begin{tabular}{l|c|c}
         & Joint Angles & Control Frequency (kHz) \\ \hline 
Baseline & 0.44 &  0.4 \\
Ours & \textbf{0.21} & \textbf{1.5}  \\
\end{tabular}
\label{table:quantitative_eval}
\end{table}

\subsection{Qualitative evaluation}
\begin{figure}[]
    \centering
    \includegraphics[width=0.48\textwidth]{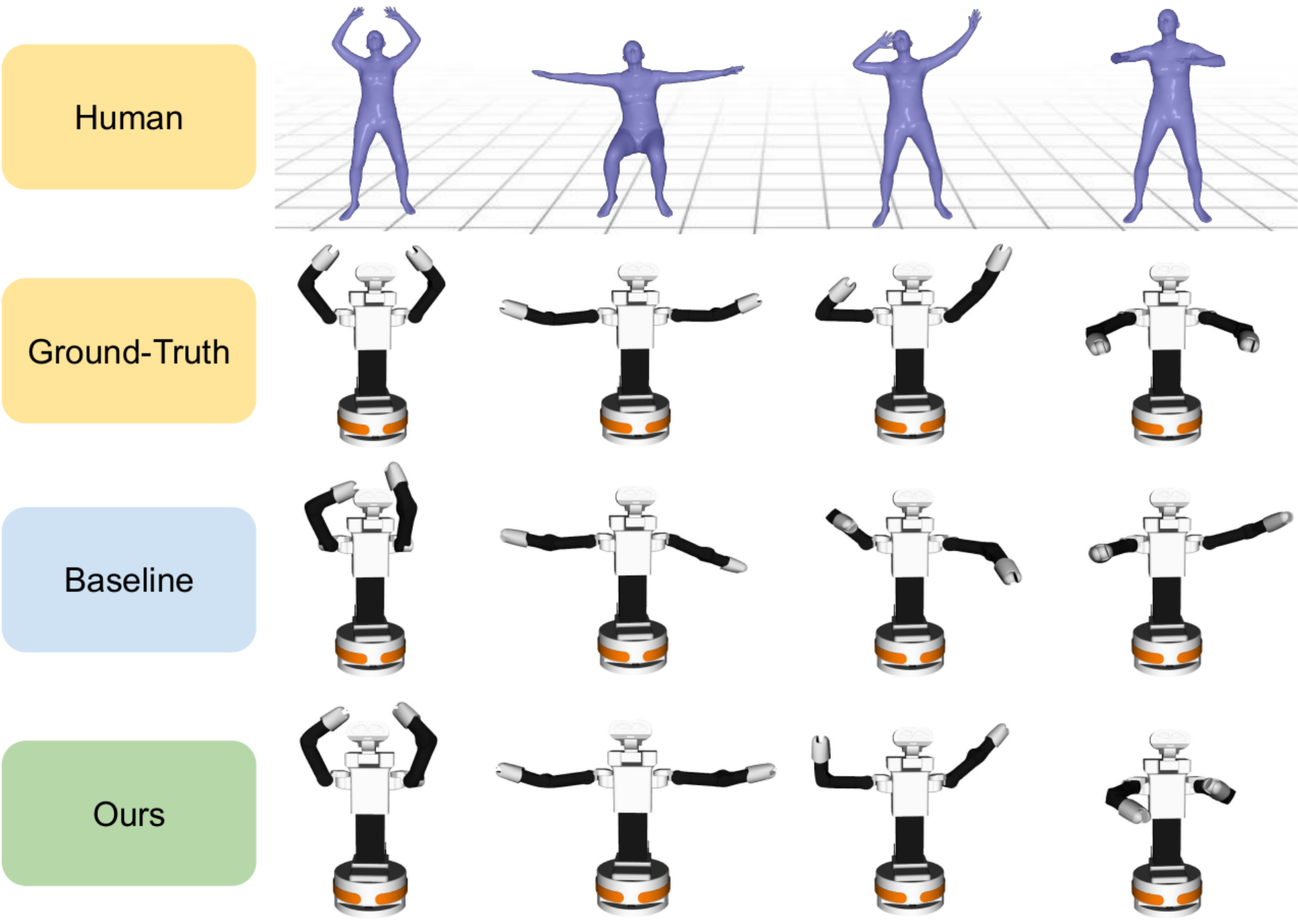}
    \caption{\textbf{Human Retargeting comparison for different key poses.} Various human skeleton key poses are retargeted to the Thiago robot. Our model captures the initial pose's visual similarity and is closely related to the manually annotated ground-truth poses.}
    \label{fig:comparison}
\end{figure}

Visually compelling examples and comparisons between the original human and retargeted robot motions are showcased in Figure \ref{fig:comparison}. For the selected human motions, we annotated their ground truth shown in the second row. Our method accurately retargets the motion when the input skeleton lifts hands above his head, lifts hands to his chest, or performs T-pose, whereas the baseline fails.

\subsection{Ablation Study}
An ablation study is conducted to systematically analyze the impact of individual loss components in our proposed model. We utilize three loss components in our approach to optimize retargeted motions. When analyzing the results in Table \ref{ablation}, it becomes apparent that the removal of the latent consistency loss $\mathcal{L}_{ltc}$ results in a slight reduction in the performance of our method. On the contrary, the Triplet loss $\mathcal{L}_{triplet}$ is indispensable for the optimization process. As supported by the experimental results, eliminating $\mathcal{L}_{triplet}$ significantly increases the loss value, rising from 0.21 to 0.57. This underscores the crucial role played by $\mathcal{L}_{triplet}$ in achieving improved optimization outcomes, contrary to all previous works that do not explore our contrastive training.

\begin{table}[]
\centering
\caption{Ablation study of proposed loss components. Mean Square Error (MSE) of joint angles between ground truth and predicted results. Bold fonts indicate better results.}
\begin{tabular}{c|c|c|c} 
$\mathcal{L}_{triplet}$ & $\mathcal{L}_{rec}$ & $\mathcal{L}_{ltc}$ & MSE \\ \hline
\cmark & \cmark & \xmark & 0.24  \\
\xmark & \cmark & \cmark & 0.57 \\
\cmark & \cmark & \cmark &  \textbf{0.21} \\
\end{tabular}
\label{ablation}
\end{table}


\subsection{From RGB videos to robot motions}
\begin{figure*}[]
    \centering
    \includegraphics[width=0.99\textwidth]{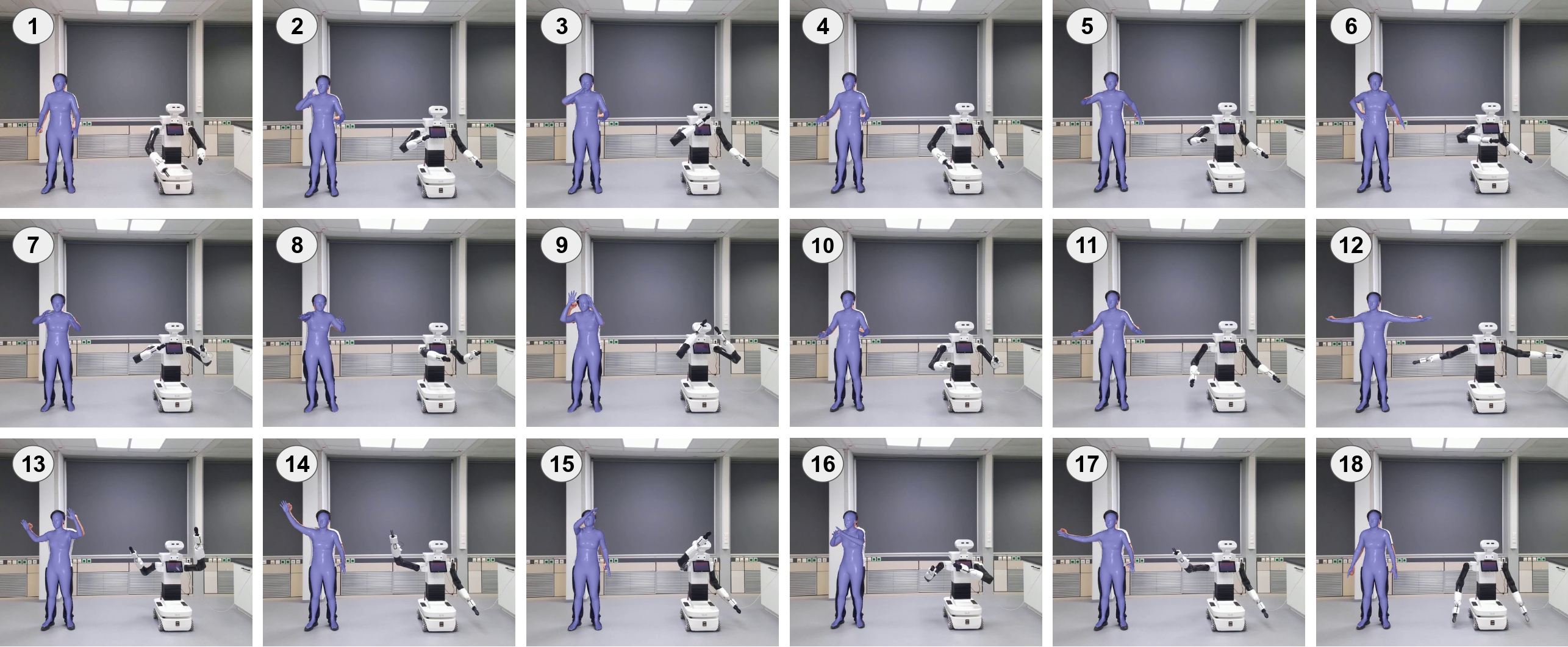}
    \caption{\textbf{Video-to-Motion.} We leverage the state-of-the-art off-the-shelf 3D human pose estimator \cite{li2021hybrik} to translate RGB images into human skeletons. Then we employ our proposed method to achieve direct motion control from human skeletons.}
    \label{fig:video-to-motion}
\end{figure*}

The proposed method can generate natural and visually similar motions from RGB videos. We adopt \cite{li2021hybrik} to obtain human 3D skeletons from RGB images in real-time. We extended \cite{li2021hybrik} with the state-of-the-art YOLOv8 \cite{YOLO} for human detection and tracking to optimize the speed. Since there is no ground truth, we only show snapshots of reference images and corresponding TiaGo poses in Figure \ref{fig:video-to-motion} for qualitative evaluation. We implement the whole pipeline that runs in real-time to control the robot's motions based on the human video.

\subsection{From texts to robot motions}
\begin{figure*}[]
    \centering
    \includegraphics[width=0.99\textwidth]{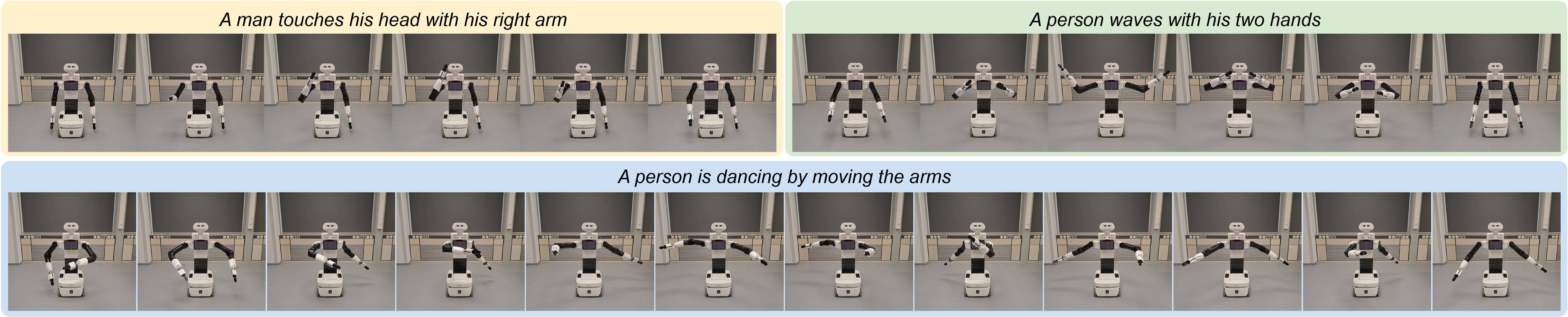}
    \caption{\textbf{Text-to-Motion.} Our model can connect as a pipeline to pre-trained motion synthesis models. In this case, we first use Text-to-Motion Retrieval \cite{petrovich23tmr} to get human motion in skeleton representation. Then, we utilize our proposed method to translate the motion into robot control commands (i.e., joint angles) to mimic it.}
    \label{fig:text-to-motion}
\end{figure*}

Text is an essential modality for human motions. Using a pre-trained motion synthesis model, Text-to-Motion Retrieval \cite{petrovich23tmr}, our method can generate robot motions with texts. To this end, we first retrieve human motions from texts with Text-to-Motion Retrieval and then retarget human motions to the TiaGo++ robot. Figure \ref{fig:text-to-motion} shows two examples of retargeting motion from texts. More examples can be found on our webpage.

\subsection{From key poses to robot motions}
\begin{figure}[]
    \centering
    \includegraphics[width=0.48\textwidth]{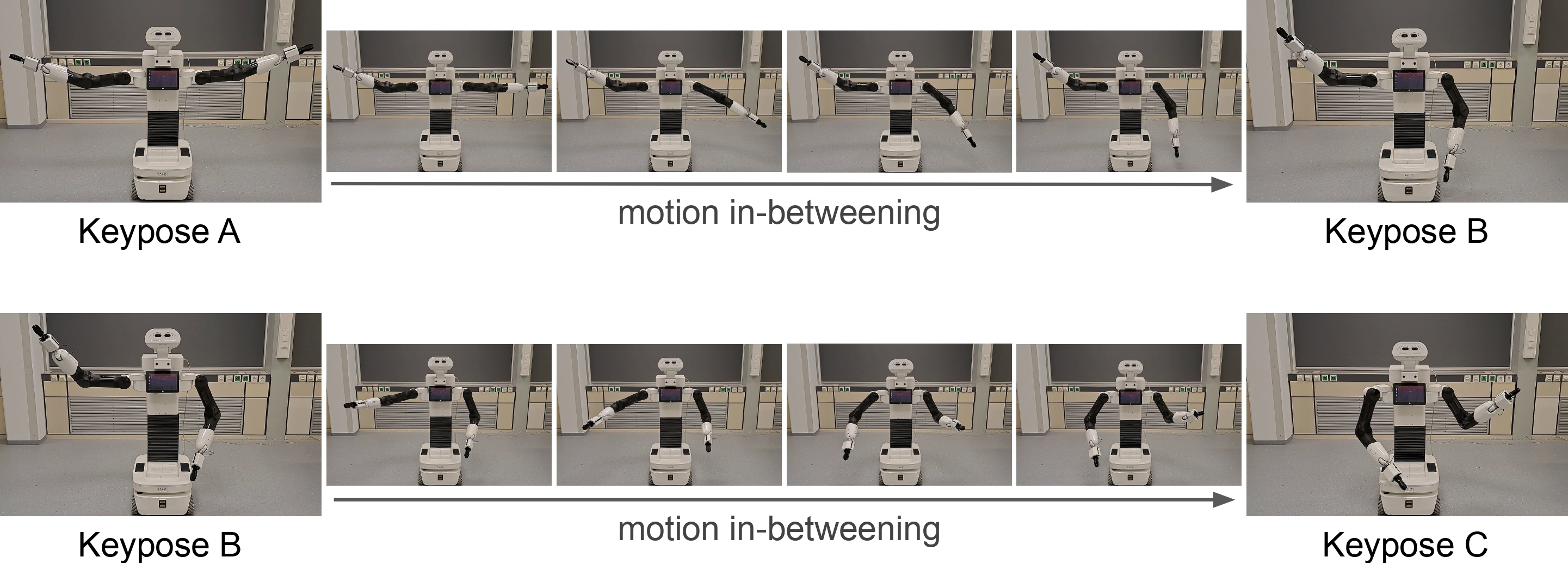}
    \caption{\textbf{Key poses-to-Motion.} The proposed method enables motion generation with key poses. Given distinct key poses, natural in-between motions can be generated by linearly interpolating key poses in our learned latent space. Our results provide the potential for direct motion control in latent space}
    \label{fig:key poses-to-motion}
\end{figure}

Our training strategy allows us to build a shared latent space that covers diverse motions. The contrastive loss $\mathcal{L}_{triplet}$ makes similar poses close and dissimilar poses far away in latent space. In such a way, our proposed method learns a smooth latent space, which enables us to interpolate motions between key poses. In Figure \ref{fig:key poses-to-motion}, we show three key poses: A, B, and C, and the interpolated in-between motions. For interpolation, two key (e.g., A and B) poses are mapped into two points in latent space, and intermediate steps can be linearly interpolated in between them. The in-between motions are decoded from these interpolated steps.

\subsection{Future work}
Our work proposed to construct a likeness-aware latent space that unifies human and robot representations seamlessly and allows for real-time robot control. While our model exhibits high precision in the retargeting process, we still observe room for improvement. Better exploring the similarity metrics between the different domains as well as connecting the shared space to higher-level representations (textual descriptions of the poses), will be considered in the future to enhance human-to-robot retargeting.

\section{CONCLUSIONS}
In this paper, we presented an unsupervised motion retargeting method that ensures a shared latent space for motion generation. To this end, we use contrastive learning combined with deep latent space modeling to incorporate human and robot motion data. To construct a shared representation of human and robot motion, we define a cross-domain similarity metric based on the global rotation of different body links. Similar motions are clustered together, and dissimilar motions are pushed apart while constructing the latent space. Furthermore, our decoder maps the shared representation to robot joint angles to control a robot directly without any additional optimization process.  Additionally, we connect our model with existing pre-trained models to achieve motion retargeting from different modalities, such as controlling the robot with given texts or retargeting from RGB videos. Moreover, our learned latent space remains tractable and allows for the generation of smooth motion inbetweening between two distinct key poses through linear interpolation in the projected latent space. We showcase all results and the robustness of our model through various experiments, both quantitatively and qualitatively.


\section*{ACKNOWLEDGMENT}
This work is funded by Marie Sklodowska-Curie Action Horizon 2020 (Grant agreement No. 955778) for project 'Personalized Robotics as Service Oriented Applications' (PERSEO).

\bibliographystyle{unsrt}
\bibliography{relatedworks}

\end{document}